%% file: main.tex
\documentclass[10pt, a4paper]{article}
\usepackage{lrec-coling2024} 
\usepackage{fitch}
\usepackage{tikz}
\newcommand{\tikzcircle}[2][color=red!40,fill=red!40]{\tikz[baseline=-0.5ex]\draw[#1,radius=#2] (0,0) circle ;}%
\usepackage{algorithm}
\usepackage{algpseudocode}
\usepackage{subcaption}
\usepackage{listings}
\usepackage{spverbatim}
\definecolor{ao(english)}{rgb}{0.0, 0.5, 0.0}
\definecolor{dkgreen}{rgb}{0,0.6,0}
\definecolor{gray}{rgb}{0.5,0.5,0.5}
\definecolor{mauve}{rgb}{0.58,0,0.82}
\definecolor{LightCyan}{rgb}{0.88,1,1}
\definecolor{bleudefrance}{rgb}{0.19, 0.55, 0.91}
\definecolor{etonblue}{rgb}{0.59, 0.78, 0.64}
\definecolor{caribbeangreen}{rgb}{0.13, 0.55, 0.13}
\usepackage{colortbl}
\usepackage{tabularx}
\newcommand*{\thead}[1]{\multicolumn{1}{c}{\bfseries #1}}
\usepackage{booktabs}
\definecolor{limegreen}{rgb}{0.2, 0.8, 0.2}
\definecolor{crimson}{rgb}{0.86, 0.08, 0.24}
\usepackage{amssymb}
\usepackage{pifont}
\newcommand{\cmark}{\ding{51}}%
\newcommand{\xmark}{\ding{55}}%
\usepackage{tcolorbox}
\usepackage{setspace}

\title{Enhancing Zero-Shot Chain-of-Thought Reasoning in Large Language Models through Logic}

\name{Xufeng Zhao, Mengdi Li, Wenhao Lu, Cornelius Weber,\\
{\bf \large 
Jae Hee Lee, Kun Chu, and Stefan Wermter} }

\address{University of Hamburg \\
         \{xufeng.zhao, wenhao.lu, cornelius.weber, jae.hee.lee, kun.chu, \\
         stefan.wermter\}@uni-hamburg.de, mengdi.li@studium.uni-hamburg.de}

\input{sections/abstract}

\begin{document}

\maketitleabstract

\input{sections/introduction}
\input{sections/relatedwork}
\input{sections/methodology}
\input{sections/experiments}
\input{sections/conclusion}
\input{sections/limitations}

\input{sections/ethics}

\section{Bibliographical References}
\bibliographystyle{lrec-coling2024-natbib}
\bibliography{custom, LoT}

\section{Language Resource References}
\bibliographystylelanguageresource{lrec-coling2024-natbib}
\bibliographylanguageresource{data}

\clearpage
\appendix
\input{sections/appendix}
\end{document}

%% file: sections/abstract.tex
\abstract{Recent advancements in large language models have showcased their remarkable generalizability across various domains. However, their reasoning abilities still have significant room for improvement, especially when confronted with scenarios requiring multi-step reasoning.
Although large language models possess extensive knowledge, their reasoning often fails to effectively utilize this knowledge to establish a coherent thinking paradigm.
These models sometimes show hallucinations as their reasoning procedures are unconstrained by logical principles.
Aiming at improving the zero-shot chain-of-thought reasoning ability of large language models, we propose \textbf{LoT} (\textbf{Lo}gical \textbf{T}houghts), a self-improvement prompting framework that leverages principles rooted in symbolic logic, particularly Reductio ad Absurdum, to systematically verify and rectify the reasoning processes step by step.
Experimental evaluations conducted on language tasks in diverse domains, including arithmetic, commonsense, symbolic, causal inference, and social problems, demonstrate the efficacy of enhanced reasoning by logic.
The implementation code for LoT can be accessed at: \url{https://github.com/xf-zhao/LoT}.
\\ \newline \Keywords{Large Language Models, Chain-of-Thought, Prompting, Reasoning, Reductio ad Absurdum}}

%% file: sections/introduction.tex
\section{Introduction}
\begin{figure}[b!]
    \centering
    \includegraphics[scale=0.7,page=1]{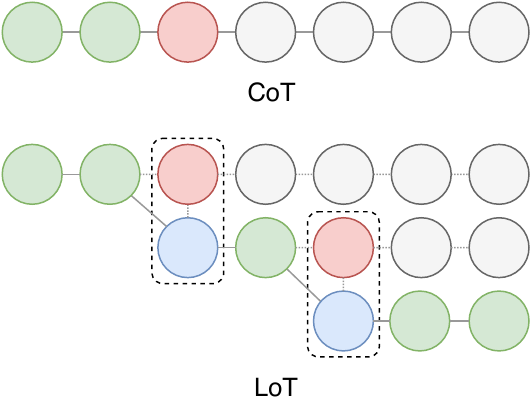}
    \caption{An overview of CoT (chain-of-thought prompting, \citealp{Wei22ChainofThoughtPrompting}) and LoT (ours). In CoT, the failure of entailment (\tikzcircle{2.5pt}) makes the rest of the deduction untrustworthy (\tikzcircle[color=gray!25, fill=gray!25]{2.5pt}), impeding the overall success of the deduction.
    In contrast, LoT is designed to think-verify-revise: it adopts those who pass the verification (\tikzcircle[color=etonblue!80, fill=etonblue!80]{2.5pt}) and revises (\tikzcircle[color=bleudefrance!55, fill=bleudefrance!55]{2.5pt}) those who do not, thereby effectively improving the overall reasoning capability.}
    \label{fig:overview}
\end{figure}
Large language models (LLMs) appear to be nearly omniscient because of their extraordinary ability to deal with tasks requiring knowledge of common sense or specialized domain knowledge.
Their success has been established in numerous fields extending beyond the realm of language processing \cite{Bubeck23SparksArtificial,Yao23ReActSynergizing,Ahn22CanNot,Zhao23ChatEnvironment}.
However, one major problem residing in generative LLMs yet to be solved is their tendency to hallucinate wrong statements in a confident style \citep{Bang23MultitaskMultilingual}. A quick example can be found by asking a non-internet-based LLM about very recent news --- it will too easily make up facts without hesitation. 

An educated human with expertise in logical reasoning can systematically examine words before coming to a conclusion.
Unlike logical reasoning by humans, the logical incompetence of deductions by LLMs makes their decisions untrustworthy.
LLMs may have a large number of logical concepts and tricks available but fail to actively utilize them in an organized manner.
However, principles in logic well-adapted by humans can leverage the reasoning ability of language models.
Take a simple logic reasoning task as an example:
``\textit{If Tom plays football outside, then John will also join to play; if John plays football, then Mary won't go outside. Knowing that Mary is outside, is Tom playing football?}'' Nine out of ten answers from ChatGPT\footnote{\url{https://openai.com/blog/chatgpt}} will conclude that ``\textit{we cannot conclude whether Tom is playing football or not}''. However, with the help of the knowledge that \textit{the contrapositive holds the exact same truth value with the original proposition}, we may prompt ChatGPT to ``use contrapositive'' to reason in another way. Then it deduces correctly: ``\textit{
... Using the contrapositive of the first statement, if John does not join to play (which we have deduced), then it implies that Tom does not play football outside. Therefore, based on the given information and the contrapositives, it can be deduced that Tom is not playing football.}''
Even though logical concepts are not new to an LLM, the model initially struggles to incorporate them.
See Fig.~\ref{fig:controposition} in Appendix~\ref{sec:appendix_chatgpt} for the full conversation.

By prompting an LLM to utilize \textit{logical equivalence}, i.e.\ being equivalent in logic
but expressed differently in natural language, 
the original premises could be represented in variational ways,
effectively fostering the exploration of potential solutions.

Motivated by the reasoning process in logic, we propose \textbf{LoT} (\textbf{Lo}gical \textbf{T}houghts) prompting framework, a fully automatic reasoning paradigm, to self-improve further the zero-shot reasoning\footnote{
Under the setting that no exemplars are provided in the prompts for in-context learning.
} ability of LLMs, which not only lets an LLM \textit{think step by step} but also verify, step by step, according to the guidance via the principle of \textit{Reductio ad Absurdum}, and revise the reasoning chain if necessary to guarantee a sound inference (see Fig.~\ref{fig:overview} for an overview).


%% file: sections/relatedwork.tex
\section{Related Work}
In order to unleash the power of a pre-trained generative language model, 
the quality of the interaction prompts plays an important role.

\subsection{Chain-of-Thought Prompting}
Prior works show that LLMs have the ability to solve complex tasks but require a proper strategy to unleash this ability, e.g.,\ human-in-the-loop alignment tuning \cite{Ouyang22TrainingLanguage} and Chain-of-Thought prompting (CoT) \cite{Wei22ChainofThoughtPrompting}.
To generate a chain of thought that decomposes the original problem into several small parts that a language model can easily handle, CoT creates few-shot exemplars of
a detailed reasoning path to let the model follow.
Least-to-most \cite{Zhou23LeasttoMostPrompting} explicitly prompts the LLM to divide complex questions into sub-problems and solve them one by one.
Moreover, zero-shot-CoT \cite{Kojima22LargeLanguage} showcases the impressive effectiveness of simply attaching the sentence ``Let's think step by step.'' before any reasoning trace starts.

We build our approach under a zero-shot setting and integrate zero-shot-CoT as a baseline to compare against.
While existing CoT-based methods focus on encouraging the reasoning step to be concrete but lack supervision of their faithfulness, we propose a step-by-step verification mechanism.

\subsection{Variational Prompting}
As an auto-regressive model, the output of an LLM can be influenced by its input.
Therefore, there are many research endeavors on prompt variations.
Summarizing existing works, the reasoning procedure benefits from prompts that (1) are \textit{relevant} to the reasoning task, (2) are \textit{diverse} in expression, (3) lead to \textit{decomposition} of complex tasks, (4) suggest \textit{grounding} with known facts, and (5) results in progressive \textit{revision} of reasoning steps.
In the design of LoT prompting, we selectively adopt these effective prompt properties.

\noindent\textbf{Relevance.}
  An LLM can be easily distracted by irrelevant words in the prompt. A pre-selection of context enhances the correctness of reasoning \cite{Creswell23SelectionInferenceExploiting,Creswell22FaithfulReasoning,Ling23DeductiveVerification}.
  Previous works typically resort to a language model to evaluate the relevance of facts and infer with the ones that contribute to a reasoning step \cite{Creswell23SelectionInferenceExploiting, Ling23DeductiveVerification}.
  Our verification of each reasoning step is conducted by prompting a language model to select relevant premises to deduct from.

\noindent\textbf{Diversity.}
  The collective intelligence from a set of reasoning paths (typically, sampling $N$ times) helps produce a reliable answer that is nearly consistent among these variants. Despite the $N$-times increased cost, this ensemble approach has been widely combined with other techniques for higher accuracy \cite{Li23MakingLarge,Ling23DeductiveVerification,Yao23TreeThoughts,Zheng23ProgressivehintPrompting}.
  A single reasoning trace may be biased.
  In order to produce a set of reasoning candidates, previous works resort to generating samples several times with the same prompt \cite{Wang23SelfConsistencyImproves}, or create diverse prompts in the beginning for variants \cite{Li23MakingLarge}.
  However, the ensemble-based approaches are both costly and inefficient. The performance of their majority voting strategy can be limited since it is not a guided, in-depth thinking strategy.

\noindent\textbf{Decomposition.}
  An automatic decomposition of a tough question improves the reasoning reliability, which has been evidenced by the success of Least-to-Most \cite{Zhou23LeasttoMostPrompting}, Zero-shot-CoT \cite{Kojima22LargeLanguage} and many prompting techniques \cite{Yao23TreeThoughts,Kojima22LargeLanguage,Wei22ChainofThoughtPrompting}.

\noindent\textbf{Grounding.}
    External functions, e.g.,\ a third-party calculator for mathematical problems \cite{Schick23ToolformerLanguage}, information acquisition from Wikipedia \cite{Yao23ReActSynergizing},  or an affordance evaluation in robotics \cite{Ahn22CanNot}, can ground the generation to be meaningful. This verification can be triggered under a specified condition or be applied to the reasoning process \cite{Lightman24LetVerify,Ling23DeductiveVerification,Li23MakingLarge}.
    LoT is primarily inspired by a logical standpoint to ground LLM generations with logical principles, empowering an LLM to argue different possibilities.
    It suggests verification and also introduces revisions of the suspected reasoning steps.

\noindent\textbf{Revision.}
        Revision (or refinement) can be regarded as a special kind of \textit{diversity}, but it is conditioned on the previous generation as hints. It re-examines the words with a focus on their quality in terms of, for example, validity and conciseness \cite{Madaan23SelfrefineIterative,Zheng23ProgressivehintPrompting,Welleck23GeneratingSequences}.
        It is an iterative generating process conditioned on previous content. Many previous works actually benefit from this manner though not explicitly mentioned. For example, Progressive-Hint Prompting \cite{Zheng23ProgressivehintPrompting} generates consistent answers by progressively guiding the LLM with hints of accumulated possible answers. It repeats the generation until the answer is deemed consistent with the previous. Other works generate content conditioned not only on the previous content but also on extra feedback \cite{Madaan23SelfrefineIterative}. To obtain a revision with high quality, this guiding feedback should be specific and actionable.
        LoT avoids unnecessary duplicating on non-controversial reasoning steps and only revises steps deemed implausible, resulting in a chain that grows only when required (Fig.~\ref{fig:overview} blue circle).
        Besides, we employ a post hoc explanation \cite{Jung22MaieuticPrompting} to provide constructive suggestions for purposeful revisions.

\subsection{Neurosymbolic Models}
Neurosymbolic models combine neural networks with symbolic representations and reasoning techniques. Their success stems from their ability to leverage symbolic (structured) knowledge to enhance learning or reasoning \cite{sarker_neuro-symbolic_2021,garcez_neurosymbolic_2020,Nye21ImprovingCoherence}. 
Unlike end-to-end black-box frameworks, these neurosymbolic models are more interpretable and explainable because of their transparency.

There exist works that adopt concepts from symbolic logic \cite{Agler12SymbolicLogic} to establish a reliable reasoning path \cite{Creswell23SelectionInferenceExploiting,Jung22MaieuticPrompting}.
To solve binary question-answering problems, \citet{Jung22MaieuticPrompting} propose to generate a post hoc explanation graph for a statement and compute the relative relations to formulate a symbolic logic expression. The truth of the statement is thereby assigned by solving the satisfiability problem of this symbolic expression.
The LoT framework employs a controlled prompting strategy that leverages logic rules and post hoc arguments to enhance error detection.


%% file: sections/methodology.tex
\vspace{-1em}
\section{Method}\label{sec:methodology}

As demonstrated in the contraposition example presented in the introduction, when known logical rules are utilized to achieve a \textit{logical equivalence}, the resultant distinct natural language expression affords LLMs a chance to engage in reasoning from an alternative perspective.

\begin{figure}[!ht]
    \centering
    \includegraphics[scale=0.8]{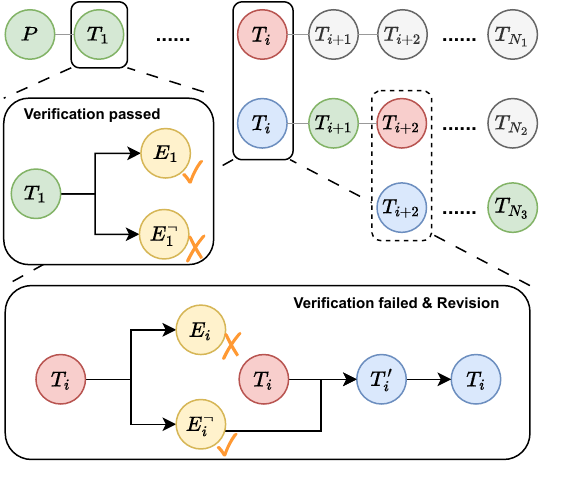}
    \caption{A diagram demonstrating the \textit{think-verify-revision} loop of LoT. The two zoom-in boxes display the processes when a thought passes (top-left) and fails (bottom) the verification respectively. A thought passing the verification is kept in the reasoning trace, while a thought failing the verification is revised and a new chain of thought is generated based on the revision. The symbols in this figure are introduced in Sec.~\ref{sec:LoT} and Sec.~\ref{sec:chain-growth}. See also Fig.~\ref{fig:example} in Appendix~\ref{sec:appendix_detail} with extended details.}
    \label{fig:explanation}
\end{figure}

A challenge is that the language model has to identify the inherent logical structures first to know whether certain prior knowledge can be effectively applied. Moreover, transforming everything from the real world into a symbolic expression is unrealistic. The applicable scenario is limited because questions in many reasoning fields beyond logic, e.g.,\ mathematics problem solving, can hardly be expressed in symbolic logic.
Nevertheless, there is promise in incorporating concepts from logic that contribute to the process of argument proof in order to construct a neurosymbolic framework \cite{garcez_neurosymbolic_2020,Creswell23SelectionInferenceExploiting} that facilitates a \textit{causal} reasoning trace, i.e.\ the premises and leading thoughts entail the thoughts behind.
Continuing with the success of ``let the model talk'', e.g.,\ ``let's think step by step'' in zero-shot-CoT \cite{Kojima22LargeLanguage}, we further propose to guide the conversation with logic for exploration of solutions. See Fig.~\ref{fig:explanation} for the guiding diagram.
\vspace{-1em}
\subsection{Reductio ad Absurdum}
Self-checking is a challenging task for LLMs \citep{Ling23DeductiveVerification,Huang24LargeLanguage}, and humans may also struggle with it.
In logic, an effective technique to establish a claim is known as \textit{\textbf{reductio ad absurdum}}, which involves an initial assumption and consequent derivation of absurdity or contradiction.

Let $P$ and $Q$ denote two propositions. The relation between a premise $P$ and its conclusion $Q$ can be expressed as $P \vdash Q$. Here ``$\vdash$'' is a syntactic turnstile which means $Q$ is a syntactic consequence of $P$ \cite{Agler12SymbolicLogic}, i.e.\ there exists a proof that claims the conclusion $Q$ given the premise $P$.
In order to prove $Q$ using \textit{reductio ad absurdum}, let us assume its negation $\neg Q$ is valid and then check the contradiction\footnote{A proposition is considered contradictory if and only if it is false under every valuation.} of the \textit{conjunctive proposition}
\begin{equation}
\label{eq:reductio}
  C = P \wedge \neg Q,
\end{equation}
where ``$\wedge$'' is a binary \textit{conjunction operator}, meaning the truth of the conjunction requires the truth of both sides. 
Upon the contradiction of the co-existence of the $P$ and $\neg Q$, $P \vdash Q$ is thus proved true, and then we can claim the validation of the conclusion $Q$ given the premise $P$.

Many logic principles, e.g.,\ the contraposition mentioned in the introduction section (see Appendix~\ref{sec:appendix_proof} for a proof), can be derived by deductions following this rule.
This thinking paradigm helps humans check arguments carefully before composing a conclusion. 
As we will demonstrate later, the reasoning ability of LLMs can also be improved by benefiting from this paradigm.

\subsection{LoT Prompting} \label{sec:LoT}
There is evidence that a series of coherent explanations helps an LLM to unleash its reasoning ability \cite{Wei22ChainofThoughtPrompting,Kojima22LargeLanguage,Zhou23LeasttoMostPrompting}, while discouragement on its utterance, e.g.,\ prompts like ``just tell me the result without any explanation'', negatively impact on an LM's reasoning ability.
So we elaborate on the success of an explicit reasoning process.

A typical $N$-step reasoning trace can be expressed as $\{ P, T_1, \cdots, T_N \}$, where $P$ is the known premise and $T_i$ is the $i$-th step of thoughts that originates from the output of a vanilla CoT.\footnote{In practice, we employ the prompt ``Let's think step by step.\textbackslash n \# 1.'' to guide an LLM to generate reasoning steps in a consistent format, each leading with a ``\# NUM. ''. Subsequently, regular expression rules can be easily applied to split these steps into atomic units $T_{\text{NUM}}$.}
Usually, $T_N$ concludes the thoughts and answers the specified question.

Unfortunately, LLMs hallucinate.
LLMs usually generate content autoregressively, which means the generation of $T_i$ is based on the former content $\{ P, \cdots, T_{i-1} \}$.
Errors in $T_i$ will propagate and gradually influence $T_{i'}$ for increasing $i' > i$, making the successive deductions and ultimately the final conclusion untrustworthy (cf. Fig.~\ref{fig:overview}).
Therefore, we propose a verification loop to double-check each reasoning step.
Following Eq.~\ref{eq:reductio}, this double-check procedure unrolls by checking the validity of $P, \cdots, T_{i-1} \vdash T_i$, i.e.\ the contradiction of
\begin{equation}
\label{eq:ci}
C_i = P \wedge T_1 \wedge \cdots \wedge T_{i-1} \wedge \mathbf{\neg} T_i,
\end{equation}
once $T_{< i}$ passed the verification. 
If any step $T_i$ fails the verification, this implies that the premises and previously verified thoughts $T_{< i}$ do not entail $T_i$. In this case, $T_{\ge i}$ needs to be revised.

To negate $T_i$ by an LLM, a straightforward way is to format $\neg T_i$ as \textit{``It is false to say $T_i$''} or to give to the LLM an instruction of \textit{``Negate $T_i$''}.
Then, the LLM has to further identify possible contradictions in $C_i$ (Eq.~\ref{eq:ci}).

\noindent\textbf{Cmps-LoT.}
Given that $T_i$ is articulated in natural language and can span multiple sentences, we aim to derive a more insightful negation by adopting the chain-of-thought methodology. Here, we task the model with \emph{composing} a post hoc explanation\footnote{A post hoc explanation is an explanation completed by the LLM with a prompt like ``$T_i$ is true because'' or ``$T_i$ is false because''.},
$E_i^{\neg}$ of $\neg T_i$~\cite{Jung22MaieuticPrompting}, and then prompt the LLM to check the validity of $C_i^{'} = C_i \wedge E_i^{\neg}$ instead. We call this simple approach Cmps-LoT.

\noindent\textbf{Adpt-LoT.}
Considering that a logical error in a text generated by an LLM is hard to spot by the LLM itself \citep{Huang24LargeLanguage}, we additionally propose to alleviate the difficulty in verifying $T_i$ by generating a pair of post hoc explanations $E_i$ and $E_i^{\neg}$ of $T_i$ and $\neg T_i$ respectively, and let the LLM decide between $T_i \wedge E_i$ and $\neg T_i \wedge E_i^{\neg}$ and \emph{adopt} one of the two.
We call this full approach Adpt-LoT or \textit{just LoT}.

An LLM is then often biased by the prompt and, as a result, generates an explanation consistent with the prompt. Because of this ``compulsory'' behavior, once a statement is deemed \textit{false} in the leading prompt, the LLM tries hard to discover errors even if they are less obvious.
LoT gains advantages from the mandatory error-detection behavior. Additionally, when transitioning from the Cmps- to the Adpt- variant, the problem transforms into a preference discrimination task \citep{Saunders22SelfcritiquingModels}, featuring more balanced reviews for both $T$ and $\neg T$ and being more manageable.
    \begin{algorithm}[ht!]
    \setstretch{1.15}
    \caption{Adpt-LoT Reasoning}
    \label{alg:logithoughts_adp}
    \begin{algorithmic}
    \Require $P$, LLM
    \State Initialize $\mathcal{T} \gets \{P\}$
    \State $T_1, T_2, \cdots, T_N \gets \text{LLM}(\mathcal{T})$ \Comment{Vanilla CoT}
    \State $i \gets 1$
    \While{$i \leq N$}
    \State $E_i^{\neg} \gets \text{post hoc LLM}(E | \neg T_i ; \mathcal{T})$
    \State \textcolor{black}{
        $E_i \gets \text{post hoc LLM}(E |T_i; \mathcal{T})$
    }
    \State \textcolor{black}{
        $\hat{E} \gets \text{LLM}(E_i; E_i^{\neg} | \mathcal{T})$ \Comment{Adopt}
    }
    \If{ \textcolor{black}{
        $\hat{E}$ is $E_i^{\neg}$
    }
        }
        \State $T_i^{'} \gets \text{LLM}(T | \mathcal{T}; T_i; E_i^{\neg})$ \Comment{Revise}
        \State $T_i \gets T_i^{'}$
        \State $\{T_{>i}\}_{N^{'}} \gets \text{LLM}(\mathcal{T} \cup T_i)$ \Comment{Adapt}
        \State $N \gets N^{'}$
    \EndIf
    \State $\mathcal{T} \gets \mathcal{T} \cup {T_i}$ \Comment{Update}
    \State $i \gets i+1$
    \EndWhile
    \State \Return $\mathcal{T}$
    \end{algorithmic}
    \end{algorithm}

\subsection{Chain Growth} \label{sec:chain-growth}

In order to investigate a step $T_i$, LoT drops all of the trailing thoughts $T_{>i}$ and branches out for revision $T_{i}^{'}$ conditioned on $\{T_{\le i}, E_i^{\neg}\}$. 
Since precise feedback is important to the success of revision \cite{Madaan23SelfrefineIterative}, we also encourage the LLM to revise any inappropriate thought with the advice to reason ``why it is wrong'', i.e.\ $E_i^{\neg}$.
Then, an adapted chain with a new conclusion can be re-generated based on the concatenation of the verified thoughts so far, i.e.\ $\{T_{<i}, T_i^{'}\}$.
This loop continues until the final conclusion passes the verification, which results in a chain with all the nodes being verified (see Fig.~\ref{fig:explanation} for an abstract depiction and Fig.~\ref{fig:example} for a detailed example). Note that this chain grows only when required.

Alg.~\ref{alg:logithoughts_adp} shows the pseudo-code of the function to compute the reasoning trace of (Adpt-)LoT. The pseudo-code for Cmps-LoT can be found in Alg.~\ref{alg:logithoughts_negation}, Appendix~\ref{sec:pseudo-code}, where the distinct procedure for identifying contradictions is emphasized.

%% file: sections/experiments.tex
\section{Experiments}
\begin{table*}[t!]
\centering
\footnotesize \setlength{\tabcolsep}{3pt}
\begin{tabular}{lccccccccc}
  \toprule
\thead{} & \thead{LoT} & \thead{GSM8K} & \thead{AQuA} & \thead{Date} & \thead{SocialQA} & \thead{Cau.Eff.} & \thead{Objects} & \thead{Letter} & \thead{OddOut}\\[0pt]
 \arrayrulecolor{gray!50} \midrule

 & \xmark & 17.52 & 21.65 & 7.24 & 37.00 & 52.94 & 34.00 & 0.00 & 25.58\\[0pt]
Vicuna-7b & \cmark & 17.68 & 20.47 & 7.24 & 36.50 & 52.94 & 35.00 & 0.00 & 25.58\\[0pt]
 &  & \textcolor{limegreen}{(+0.16)} & \textcolor{crimson}{(-1.18)} & (0.00) & \textcolor{crimson}{(-0.50)} & (0.00) & \textcolor{limegreen}{(+1.00)} & (0.00) & (0.00)\\[0pt]

 \arrayrulecolor{gray!20}\hline
  \rowcolor{gray!10}
 & \xmark & 33.79 & 22.05 & 32.31 & 41.00 & 68.75 & 31.00 & 2.00 & 29.07\\[0pt]
  \rowcolor{gray!10}
Vicuna-13b & \cmark & 37.56 & 23.62 & 33.15 & 48.50 & 68.75 & 31.50 & 4.00 & 45.35 \\[0pt]
  \rowcolor{gray!10}
 &  & \textcolor{limegreen}{(+3.77)} & \textcolor{limegreen}{(+1.57)} & \textcolor{limegreen}{(+0.84)} & \textcolor{limegreen}{(+7.50)} & (0.00) & \textcolor{limegreen}{(+0.50)} & \textcolor{limegreen}{(+2.00)} & \textcolor{limegreen}{(+16.28)}\\[0pt]
 \arrayrulecolor{gray!20}\hline

 & \xmark & 40.33 & 26.38 & 15.70 & 37.50 & 52.94 & 32.00 & 14.67 & 40.70\\[0pt]
Vicuna-33b & \cmark & 40.49 & 29.53 & 20.35 & 47.50 & 68.75 & 34.50 & 14.00 & 43.02\\[0pt]
 &  & \textcolor{limegreen}{(+0.16)} & \textcolor{limegreen}{(+3.15)} & \textcolor{limegreen}{(+4.65)} & \textcolor{limegreen}{(+10.00)} & \textcolor{limegreen}{(+15.81)} & \textcolor{limegreen}{(+2.50)} & \textcolor{crimson}{(-0.67)} & \textcolor{limegreen}{(+2.32)}\\[0pt]

 \arrayrulecolor{gray!20}\hline
  \rowcolor{gray!10}
 & \xmark & 78.75 & 57.09 & 51.26 & 72.00 & 92.16 & 60.75 & 67.33 & 81.40\\[0pt]
  \rowcolor{gray!10}
GPT-3.5-turbo & \cmark & 80.15 & 60.63 & 52.37 & 72.00 & 92.16 & 58.25 & 67.33 & 81.40\\[0pt]
  \rowcolor{gray!10}
 &  & \textcolor{limegreen}{(+1.40)} & \textcolor{limegreen}{(+3.54)} & \textcolor{limegreen}{(+1.11)} & (0.00) & (0.00) & \textcolor{crimson}{(-2.50)} & (0.00) & (0.00)\\[0pt]
 \arrayrulecolor{gray!20}\hline

 & \xmark & 94.29 & 71.56 & 83.09 & 77.50 & 100.00 & 100.00 & 92.61 & 95.35\\[0pt]
GPT-4 & \cmark & 95.71 & 74.31 & 85.16 & 77.50 & 100.00 & 100.00 & 93.14 & 96.51\\[0pt]
 &  & \textcolor{limegreen}{(+1.42)} & \textcolor{limegreen}{(+2.75)} & \textcolor{limegreen}{(+2.07)} & (0.00) & (0.00) & (0.00) & \textcolor{limegreen}{(+0.53)} & \textcolor{limegreen}{(+1.16)}\\[0pt]

 \arrayrulecolor{gray} \bottomrule
\end{tabular}
\caption{
We evaluate the accuracy of our methods and compare them against baseline approaches using various models and datasets, with computation based on ground truth annotations.
The percentage difference of CoT \cite{Kojima22LargeLanguage} without (\xmark) and with (\cmark) LoT enhancement using different LLMs is shown below each cell group (green if positive, red if negative). CoT generally gains better performance when being enhanced by LoT. Larger models, such as GPT-4, exhibit strong and robust self-correction ability.}
\label{tb:result}
\end{table*}
For the following reasons we carry out the experiments in a zero-shot setting:
1) Zero-shot-CoT \cite{Kojima22LargeLanguage} has a wide task-agnostic application potential, while few-shot requires domain knowledge;
2) The few-shot prompts heavily influence the performance even on the same dataset, so it is hard to evaluate fairly as the prompt varies.
Drawing direct comparisons with other prompting works in the literature is challenging due to variations in task settings and backend language models. Many of these works are specifically under a few-shot setting, which would necessitate additional modifications to adapt them for zero-shot reasoning.
We consider this as an area for future investigation.

Since our work is an enhancement on the chain produced by zero-shot-CoT \cite{Kojima22LargeLanguage}, we compare LoT with it as the baseline to demonstrate the benefit of step-wise verification and revision for zero-shot reasoning.
We evaluate the accuracy of tasks in various domains as the overall performance measure and also report the impact of the logical revision on the original reasoning chain.

We aim to answer the following research questions by conducting experiments:
\begin{enumerate}
    \item Does LoT enhance the performance of CoT in various domains as well as with LLMs of varying model scales? (Sec.~\ref{sec:q1})
    \item What is the impact of LoT on individual reasoning chains (e.g.\ revision frequency, length)? (Sec.~\ref{sec:q2})
    \item Do post-hoc explanations help LLM self-check? (Sec.~\ref{sec:q3})
\end{enumerate}

\subsection{Experimental Setup}
\textbf{Dataset.} We demonstrate the effectiveness of LoT on diverse language topics:

    (1) Math reasoning tasks GSM8K \citetlanguageresource{Cobbe21TrainingVerifiers} and AQuA  \citetlanguageresource{Ling17ProgramInduction}.
    The GSM8K dataset contains grade school mathematics questions that should be responded to by numerical answers; AQuA has more advanced questions but has several optional answers to choose from.

    (2) Commonsense reasoning tasks DateUnderstanding and OddOneOut \citetlanguageresource{Srivastava23ImitationGame}.
    The DateUnderstanding task necessitates the utilization of both common sense and fundamental arithmetic calculations to find out the correct date, making it sufficiently challenging to prevent it from being solvable through simple one-step reasoning. The OddOneOut requires common sense to deduct the unusual object in the context.

    (3) Causal inference tasks CauseEffect and ShuffledObjects \citelanguageresource{Srivastava23ImitationGame}, where both of the tasks require reasoning from the context for a correct deduction.

    (4) Symbolic reasoning task LastLetter \citelanguageresource{Srivastava23ImitationGame}. In this task, the language model has to extract the last letter of given candidates and concatenate them in order, which is simple for humans but challenging for language models because of tokenization \citelanguageresource{Mielke21WordsCharacters}.

    (5) Social interaction reasoning task, SocialQA \citelanguageresource{Srivastava23ImitationGame}, that measures the model's emotional and social intelligence in human daily activities. Completing the task requires an understanding of human behavior.
    
To get a formatted answer that can be directly compared with the ground truth in the aforementioned dataset, a final prompt asking the final answer is attached after the reasoning trace, e.g.,\ for the GSM8K dataset we simply attach ``Therefore, the final numerical answer is:'' at the ends. For robustness, this answer is matched with a regular expression to only extract numerical digits before comparing it with the ground truth.

\noindent\textbf{Backend LLMs.} Previous works show that the performance improvement of the CoT technique varies when applied on language models of different capabilities \cite{Wei22ChainofThoughtPrompting,Kojima22LargeLanguage}. Therefore, we conducted an evaluation of the LoT method using a range of models, including Vicuna-7b, Vicuna-13b, and Vicuna-33b models \citep{Chiang23VicunaOpensource}, as well as GPT-3.5-turbo and GPT-4.
The Vicuna model is an open-sourced language model trained by fine-tuning LLaMA \citep{Touvron23LLaMAOpen} on user-shared conversations.
It demonstrates strong performance across various scenarios and offers flexibility in terms of model size selection.
On the other hand, GPT-3.5-turbo and GPT-4 are larger models known for their state-of-the-art performance in numerous tasks.

To ensure stable results and promote self-error detection within the models, we set the temperature parameter to 0.1. Additionally, the max\_token parameter was established at 2048, a sufficient limit to accommodate all the datasets employed in our evaluation.
\vspace{-1em}
\subsection{Does LoT enhance the performance of CoT?}\label{sec:q1}
To answer the first question, we conduct zero-shot experiments with datasets covering more diverse topics and with language models of different sizes.
The LoT-enhanced performance is compared with the zero-shot baseline in Tab.~\ref{tb:result}.
The experiment shows that LoT can enhance the performance of the base CoT in various domains. The performance benefits are more consistent when the model size gets considerable (>7B parameters).
Moreover, the performance gain becomes more prominent as the model’s ability increases (e.g. GPT-4).


\subsection{What is the impact on individual reasoning chains?}\label{sec:q2}
We report more insightful case-wise statistics and discussions in this section, including
(1) average revision frequency in Tab.~\ref{tab:revision_times};
(2) the resultant number of reasoning steps in Tab.~\ref{tab:steps};
and (3) a case study to illustrate the logical reasoning procedure.
More detailed statistics including the worsening rate (i.e. the ones being originally correct by CoT but ``corrected'' to be wrong by LoT) and improvement rate (i.e. the ones that are originally wrong and being corrected by LoT) can be found in Tab.~\ref{tab:imprworsenrate}, Appendix~\ref{sec:appendix_worsenimprove}.
\begin{table*}[t!]
\begin{center}
\footnotesize \setlength{\tabcolsep}{3pt}
\begin{tabular}{lcccccccc}
  \toprule
\thead{Revision $\circlearrowleft$} & \thead{GSM8K} & \thead{AQuA} & \thead{Date} & \thead{SocialQA} & \thead{Cau.Eff.} & \thead{Objects} & \thead{Letter} & \thead{OddOut}\\[0pt]
 \arrayrulecolor{gray!50} \midrule

Vicuna-7b & 2\% & 4\% & 2\% & 1\% & 2\% & 0\% & 3\% & 0\%\\[0pt]

 \arrayrulecolor{gray!20}\hline
  \rowcolor{gray!10}
Vicuna-13b & 7\% & 10\% & 5\% & 5\% & 0\% & 7\% & 2\% & 0\%\\[0pt]
 \arrayrulecolor{gray!20}\hline

Vicuna-33b & 2\% & 9\% & 8\% & 7\% & 6\% & 9\% & 1\% & 7\%\\[0pt]

 \arrayrulecolor{gray!20}\hline
  \rowcolor{gray!10}
GPT-3.5-turbo & 16\% & 28\% & 32\% & 5\% & 20\% & 9\% & 4\% & 16\%\\[0pt]
 \arrayrulecolor{gray!20}\hline

GPT-4 & 3\% & 20\% & 7\% & 2\% & 0\% & 1\% & 0\% & 8\%\\[0pt]

 \arrayrulecolor{gray} \bottomrule
\end{tabular}
\caption{The average step-wise revision frequency is presented as a percentage, reflecting how often a reasoning step is revised by LoT.}
\label{tab:revision_times}
\end{center}
\end{table*}

\begin{table*}[t!]
\begin{center}
\footnotesize \setlength{\tabcolsep}{3pt}
\begin{tabular}{lccccccccc}
  \toprule
\thead{} & \thead{LoT} & \thead{GSM8K} & \thead{AQuA} & \thead{Date} & \thead{SocialQA} & \thead{Cau.Eff.} & \thead{Objects} & \thead{Letter} & \thead{OddOut}\\[0pt]
 \arrayrulecolor{gray!50} \midrule

Vicuna-7b & \xmark & 1.22 & 1.16 & 1.34 & 1.09 & 1.00 & 2.54 & 3.46 & 1.00\\[0pt]
 & \cmark & 1.27 & 1.21 & 1.35 & 1.10 & 1.02 & 2.54 & 3.49 & 1.00\\[0pt]

 \arrayrulecolor{gray!20}\hline
  \rowcolor{gray!10}
Vicuna-13b & \xmark & 2.81 & 2.89 & 5.06 & 2.69 & 1.00 & 2.93 & 1.66 & 1.00\\[0pt]
  \rowcolor{gray!10}
 & \cmark & 2.74 & 2.87 & 5.05 & 2.71 & 1.00 & 2.96 & 1.69 & 1.00\\[0pt]
 \arrayrulecolor{gray!20}\hline

Vicuna-33b & \xmark & 1.94 & 1.99 & 2.31 & 3.26 & 1.00 & 3.26 & 1.20 & 1.70\\[0pt]
 & \cmark & 1.94 & 1.91 & 2.33 & 3.13 & 1.06 & 3.23 & 1.21 & 1.64\\[0pt]

 \arrayrulecolor{gray!20}\hline
  \rowcolor{gray!10}
GPT-3.5-turbo & \xmark & 4.17 & 6.83 & 3.66 & 2.50 & 1.73 & 3.02 & 4.84 & 1.57\\[0pt]
  \rowcolor{gray!10}
 & \cmark & 4.08 & 6.24 & 3.56 & 2.51 & 1.92 & 3.05 & 4.81 & 1.70\\[0pt]
 \arrayrulecolor{gray!20}\hline

GPT-4 & \xmark & 3.42 & 4.22 & 2.71 & 2.33 & 1.00 & 3.00 & 4.05 & 1.00\\[0pt]
 & \cmark & 3.41 & 4.39 & 2.72 & 2.33 & 1.00 & 3.01 & 4.05 & 1.08\\[0pt]

 \arrayrulecolor{gray} \bottomrule
\end{tabular}
\caption{The average number of resultant reasoning steps without (\xmark) and with (\cmark) LoT applied.}
\label{tab:steps}
\end{center}
\end{table*}

\noindent\textbf{Revision Frequency.}
In order to measure the complexity of revisions, we list the average revisions \textit{per chain} in Tab.~\ref{tab:revision_times} and typical reasoning steps required by CoT and LoT in Tab.~\ref{tab:steps}.
Note that the number of steps is not human-defined or prompted since our setting is in zero-shot, so the language models decide by themselves the length of a reasoning chain.

The percentage of revisions indicates the frequency of LoT to revise the candidate reasoning chain.
We observe that language models with powerful capabilities (e.g.,\ GPT-4 and GPT-3.5-turbo) are more active in revision than smaller models; and challenging tasks such as the math reasoning task lead to more revisions.
However, revision does not necessarily alter the final deduction answer. For example, LoT with GPT-3.5-turbo backend revises several times on SocialQA yet achieves the same accuracy as CoT. Intuitively, solving one problem may lead to multiple pathways, and some revisions might aim to enrich the sentence by incorporating additional known conditions and rhetorical supplements.

\noindent\textbf{Resultant Steps.}
The average step count is the valid reasoning steps in the final CoT and LoT reasoning paths (i.e. the intermediate verification, refinement, etc. are not shown).
From Tab.~\ref{tab:steps}, we can conclude that 1) larger language models generally generate longer chains and are also more active in revision; 2) The LoT refined reasoning chain is generally a little bit shorter than the original zero-shot CoT. Our conjecture is that this phenomenon might arise because, during the refinement process, the language model strives to incorporate additional information, consequently yielding concise chains of reasoning.

\noindent\textbf{Case Study.}
We show a successful case on the Date Understanding task to demonstrate the verification and revision procedure applied to the chain of thoughts initialized by zero-shot-CoT. (See Appendix~\ref{sec:appendix_cases} for detailed prompts and further case studies.)

In Snippet A, there are initialized zero-shot-CoT reasoning steps where step \#6 is actually incorrectly inferred (colored in red).
Here, \textit{black} color indicates given context or fixed prompts; \textit{non-black} color indicates generated content by the LLM.
The error occurs because zero-shot-CoT is distracted by the irrelevant premise of ``\textit{Jane's appointment will be 3 days later}'' and concludes with a wrong answer.
\begin{tcolorbox}[
    title={\footnotesize{Snippet A}},
    fonttitle=\fontsize{9}{12}\selectfont, 
    coltitle=black!95,
    colframe=gray!20,
    boxrule=0.5pt,
    standard jigsaw,
    opacityback=0,  
    boxsep=0pt,
    left=4pt,
    right=4pt,
    top=4pt,
    bottom=4pt,
    label=snipA,
    fontupper=\linespread{0.8}\selectfont,
    fontlower=\linespread{0.8}\selectfont,]
    \setstretch{.98}
\begin{footnotesize}
\textcolor{gray}{Question: Today is Apr 10, 1985. Jane's appointment will be 3 days later. What is the date 10 days ago in MM/DD/YYYY?}\\
\textcolor{gray}{Let's think step by step.} \\
\textcolor{gray}{Answer:}\\
\textcolor{caribbeangreen}{\#1. Today is Apr 10, 1985.} \\
\textcolor{caribbeangreen}{\#2. Jane's appointment will be 3 days later.} \\
\textcolor{caribbeangreen}{\#3. So, Jane's appointment will be on Apr 13, 1985.} \\
\textcolor{caribbeangreen}{\#4. To find the date 10 days ago, we subtract 10 days from Apr 13, 1985.} \\
\textcolor{caribbeangreen}{\#5. Apr 13, 1985 - 10 days = Apr 3, 1985.} \\
\textcolor{red!70}{\#6. Therefore, the date 10 days ago is Apr 3, 1985.}
\end{footnotesize}
\end{tcolorbox}
\begin{tcolorbox}[
    title={\footnotesize{Snippet B}},
    fonttitle=\fontsize{9}{12}\selectfont, 
    coltitle=black!95,
    colframe=gray!20,
    boxrule=0.5pt,
    standard jigsaw,
    opacityback=0,  
    boxsep=0pt,
    left=4pt,
    right=4pt,
    top=4pt,
    bottom=4pt,
    float=ht!,
    fontupper=\linespread{0.8}\selectfont,
    fontlower=\linespread{0.8}\selectfont,]
    \setstretch{.98}
\begin{footnotesize}
\textcolor{gray}{Verification of the next step:} \\
\textcolor{gray}{\#6. Therefore, the date 10 days ago is Apr 3, 1985.} \\
\textcolor{caribbeangreen}{Review X: \textless review\textgreater step \#6 is TRUE because we are subtracting 10 days from the given date of Apr 13, 1985. By subtracting 10 days, we go back in time and arrive at Apr 3, 1985. Therefore, the date 10 days ago is Apr 3, 1985. \textless /review\textgreater} \\
\textcolor{caribbeangreen}{Review Y: \textless review\textgreater step \#6 is FALSE because the reasoning in step \#4 is incorrect. To find the date 10 days ago, we need to subtract 10 days from Apr 10, 1985, not from Apr 13, 1985.} \\
\textcolor{caribbeangreen}{The correct calculation would be:} \\
\textcolor{caribbeangreen}{Apr 10, 1985 - 10 days = Mar 31, 1985.} \\
\textcolor{caribbeangreen}{Therefore, the date 10 days ago is Mar 31, 1985. \textless /review\textgreater}
\end{footnotesize}
\end{tcolorbox}
To verify the 6th step, as shown in Snippet~B, the LLM is prompted to generate two opposite reviews\footnote{In practice, as is shown in the example case, the special language mark ``<review></review>'' is applied for the sake of robust separations of different reviews.} given the question context and the previous five verified steps.
Due to the content limitation, we omit the long prompt and focus on the resultant generation. Note that \textit{Review Y} successfully detects the unobvious error and provides a correction that can be adopted later for revision.
In Snippet C, the LLM is encouraged to extract and deduct based on relevant premises. Finally, it ends with adopting \textit{Review Y} and reaching a correct conclusion.

\begin{tcolorbox}[
    title={\footnotesize{Snippet C}},
    fonttitle=\fontsize{9}{12}\selectfont, 
    coltitle=black!95,
    colframe=gray!20,
    boxrule=0.5pt,
    standard jigsaw,
    opacityback=0,  
    boxsep=0pt,
    left=4pt,
    right=4pt,
    top=4pt,
    bottom=4pt,
    float=t!,
    fontupper=\linespread{0.8}\selectfont,
    fontlower=\linespread{0.8}\selectfont,]
    \setstretch{.98}
\begin{footnotesize}
\textcolor{gray}{Let's check two different reviews (X and Y).}\\
\textcolor{gray}{Support the more plausible one and criticize the other one.}\\
\textcolor{caribbeangreen}{Review X: ...}\\
\textcolor{caribbeangreen}{Review Y: ...}\\
\textcolor{gray}{Let's start by analyzing one by one.}\\
\textcolor{gray}{I. What are the premises and previous steps to support the verification of step \#6? (Your answer should quote the exact quote as support.)}\\
\textcolor{gray}{II. Criticise the incorrect review.}\\
\textcolor{gray}{(Note the examined step doesn't have to tackle the whole problem at once.)}\\
\textcolor{gray}{Finally, identify whether step \#6 is true or false.}\\
\textcolor{gray}{Analysis and conclusion:}\\
\textcolor{caribbeangreen}{I. The premises and previous steps to support the verification of step \#6 are as follows:}\\
\textcolor{caribbeangreen}{- Today is Apr 10, 1985.}\\
\textcolor{caribbeangreen}{- Jane's appointment will be 3 days later.}\\
\textcolor{caribbeangreen}{- Jane's appointment will be on Apr 13, 1985.}\\
\textcolor{caribbeangreen}{- To find the date 10 days ago, we subtract 10 days from Apr 13, 1985.}\\
\textcolor{caribbeangreen}{II. Review Y is correct in criticizing step \#4. The reasoning in step \#4 is indeed incorrect. To find the date 10 days ago, we need to subtract 10 days from Apr 10, 1985, not from Apr 13, 1985.}\\
\textcolor{caribbeangreen}{The correct calculation would be:}\\
\textcolor{caribbeangreen}{Apr 10, 1985 - 10 days = Mar 31, 1985.}\\
\textcolor{caribbeangreen}{Therefore, the date 10 days ago is Mar 31, 1985.}\\
\textcolor{caribbeangreen}{Conclusion: Step \#6 is false. The correct date 10 days ago is Mar 31, 1985, not Apr 3, 1985.}
\end{footnotesize}
\end{tcolorbox}

\subsection{Do post-hoc explanations help LLM self-check?}\label{sec:q3}
In order to examine the error-finding ability, the results of using LoT and its ablated variants on three challenging tasks are shown in Tab.~\ref{tab:ablation},
where ``Self-Check'' is an ablation of \textit{reductio ad absurdum} such that the LLM is only prompted to ``double check the step'' without logical hints.

With LoT, the complex examination problem is broken down into two procedures: 1) the post hoc explanation mechanism for rigorous error detection, and 2) a single-choice preference for selecting the more reasonable opinion from the two candidates.
The better performance of (Adpt-)LoT suggests that when it comes to error detection in LLM's reasoning, it is more effective for an LLM to embrace one of two opposing viewpoints ($T, E$ or $\neg T, E^\neg$) rather than \textit{composing} the verification directly (which is the case for both ``Self-Check'' and Cmps-LoT), especially when coping with tasks that are difficult such as math reasoning.
\begin{table}[ht!]
\setstretch{1.35}
\centering
\footnotesize \setlength{\tabcolsep}{3pt}
\begin{tabularx}{0.68\linewidth}{lccc}
\toprule
\textbf{Method} & \textbf{GSM8K} & \textbf{AQuA}  & \textbf{Date}\\
 \arrayrulecolor{gray!50} \midrule

CoT & 78.75 & 57.09 & 51.26 \\

\arrayrulecolor{gray!20}\hline
\rowcolor{gray!10}
Self-Check & 76.15 & 56.19 & 51.57 \\
 \arrayrulecolor{gray!20}\hline

Cmps-LoT & 77.67 & 57.48 & \textbf{52.37} \\

\arrayrulecolor{gray!20}\hline
\rowcolor{gray!10}
LoT & \textbf{80.15} & \textbf{60.63} & \textbf{52.37} \\
 \arrayrulecolor{gray!20}\hline

 \arrayrulecolor{gray} \bottomrule
\end{tabularx}
\caption{Zero-shot accuracy results (in \%) in comparison of LoT (i.e.\ Adpt-) and its ablated variants. The underlying LLM is GPT-3.5-turbo.} \label{tab:ablation}
\end{table}

%% file: sections/conclusion.tex
\vspace{-1em}
\section{Conclusion and Future Work}
LLMs have impressive reasoning ability in domains that require commonsense knowledge, specialized expertise, comprehensive understanding, etc. However, there is still room to improve their multi-step reasoning capability.
Building upon zero-shot-CoT, we derive the LoT prompting framework from a symbolic logic perspective, utilizing the widely applicable principle of \textit{reductio ad absurdum}, resulting in a robust \textit{think-verify-revise} framework with plausible prompting properties.
Experiments conducted on a variety of reasoning tasks spanning different domains demonstrate that enhancing zero-shot Chain-of-Thought with LoT leads to improved reasoning ability, particularly when applied to LLMs of large scale.

\noindent \textbf{Future Work.}
While our research primarily focuses on integrating human knowledge into CoT prompting, further exploration of additional logical deduction principles could enhance the reasoning process. Moreover, we demonstrate the efficacy of bolstering the robustness of complex reasoning by discerning between conflicting outputs, suggesting the potential extension of this approach to prompt and refine LLMs for self-improvement. This may entail utilizing self-checked outputs for reinforcement learning from AI feedback (RLAIF) \citep{Li_Awesome_RLAIF,Lee23RLAIFScaling}. Such endeavors show promise, particularly in situations where a positive ``GD gap'' exists \citep{Saunders22SelfcritiquingModels}, i.e., language models with promises to be further improved by discerning the quality of its generation, but we defer the investigation of this avenue to future research endeavors.

%% file: sections/limitations.tex
\section*{Limitations}
\textbf{Prompting and Fine-tuning.} LoT establishes a controlled prompting strategy for self-correction.
Nonetheless, it is worthwhile to explore future endeavors of fine-tuning LLMs for the purpose of spontaneous logical reasoning.

\noindent \textbf{Generation Probability.} Rather than letting the LLM choose from different reviews, another possible method is to access and compare the probability of the generations.
Unfortunately, there is no public access to the generation probability of GPT-3.5-turbo yet\footnote{\url{https://platform.openai.com/docs/api-reference}} as it is possible for completion models (such as text-davinci-003). 
Considering a cheaper price and better performance, we conducted our experiments with the chatting model and left this possibility for future work.

\noindent \textbf{Zero-shot, Few-shot and Beyond.} 
Since our work is done with an aim to be as generalizable as possible, the experiments are all conducted in the zero-shot setting.
Nonetheless, incorporating domain knowledge into the exemplar prompt proves advantageous for enhancing performance \cite{Kojima22LargeLanguage,Wei22ChainofThoughtPrompting},
it is still worthwhile to explore the potential when LoT is applied in the few-shot setting in future work.

%% file: sections/ethics.tex
\section*{Ethics Statement}

Large language models sometimes produce biased, untrustworthy statements. Despite our best intention to enhance the model, we are not rectifying these issues. 
It is advisable for individuals to exercise caution and avoid placing excessive reliance on it.
This method is released with the purpose of research only.

\section*{Acknowledgements}
We gratefully acknowledge support from the China Scholarship Council (CSC) and the German Research Foundation (DFG) under the project Crossmodal Learning (TRR 169).

%% file: sections/appendix.tex
\onecolumn

\section{ChatGPT Deduction Example w/ and w/o Contraposition}\label{sec:appendix_chatgpt}
\begin{figure*}[ht!]
    \centering
    \includegraphics[scale=0.9]{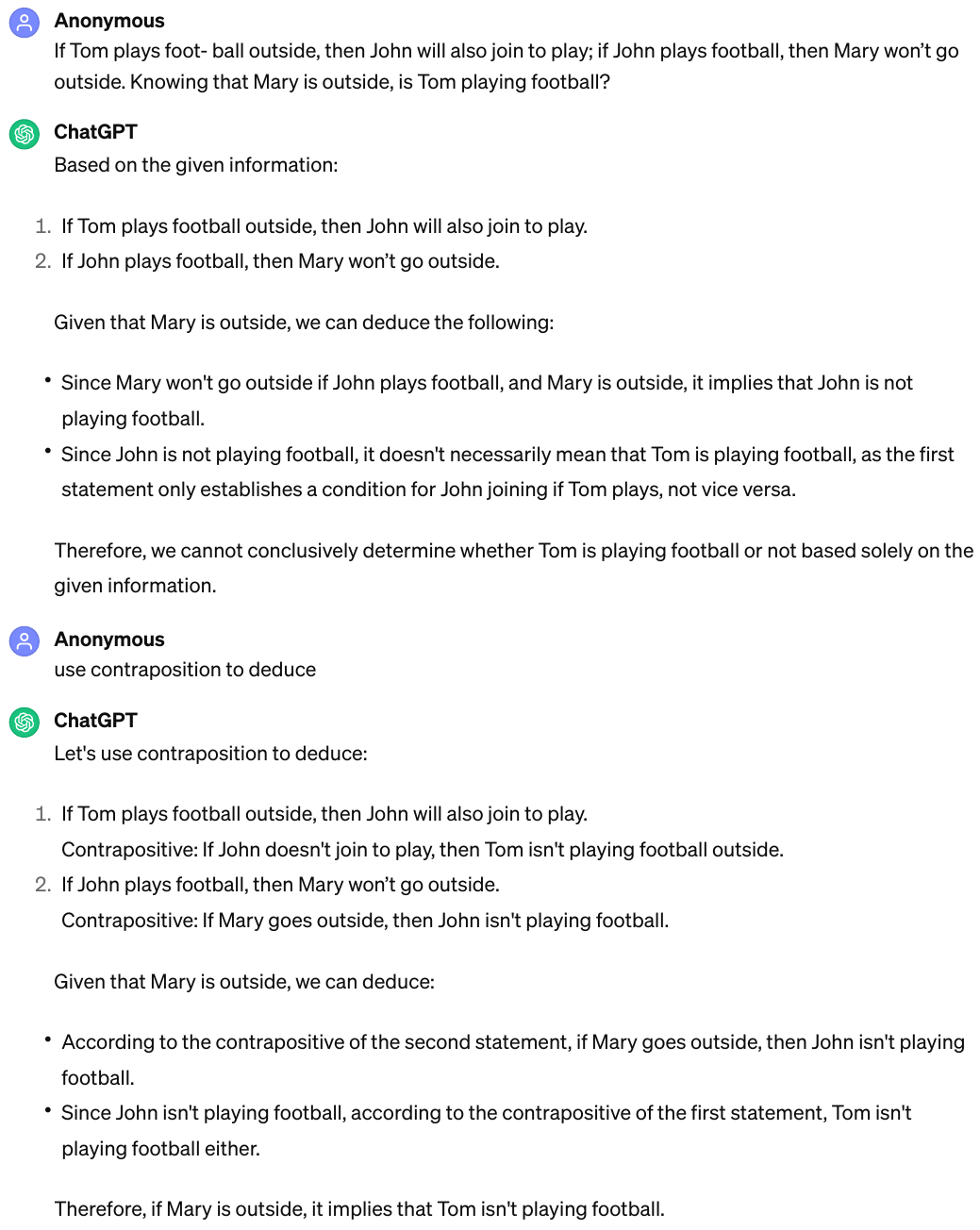}
    \caption{An example conversation with ChatGPT where the language model fails to correctly deduce the answer initially, but when being prompted to use the idea of ``contraposition'', it successfully reaches the desired result.}
    \label{fig:controposition}
\end{figure*}

Fig.~\ref{fig:controposition} shows the full ChatGPT conversation of the ``football'' example introduced earlier in the introduction section, indicating the benefits of explicitly prompting LLMs with deduction knowledge such as logic principles.

\section{Proof of Contraposition}
\label{sec:appendix_proof}

We will prove contraposition using \textit{reductio ad absurdum}:
\begin{tcolorbox}
Known premises $P \rightarrow Q$ (if $P$ is true, then $Q$ is true) and $\neg Q$ ($Q$ is false), prove $\neg P$.
\end{tcolorbox}
Proof $P \rightarrow Q, \neg Q \vdash \neg P$:
\begin{center}
\begin{fitch}
\fa P \rightarrow Q & $\mathbf{P}$ \\
\fj \neg Q          & $\mathbf{P}$ \\
\fa \fh P           & $\mathbf{A}$ \\
\fa \fa Q           & $\rightarrow \mathbf{E}$ 1, 3 \\
\fa \fa \text{X}    & $\mathbf{C}$ 2, 4 \\
\fa \neg P          & $\neg \mathbf{I}$ 3-5 \\
\end{fitch}
\end{center}

Inside the proof, $\mathbf{P}$ stands for the known premise, $\mathbf{A}$ for assumption, $\rightarrow \mathbf{E}$ for the elimination of ``$\rightarrow$'' symbol by following the conditional statement, $\mathbf{C}$ for contradiction assertion (followed by an ``X'' which indicates the branch with an assumption is \textit{closed}) and $\neg \mathbf{I}$ for the introduction of negation according to the rule of \textit{reductio ad absurdum}.

\section{Pseudo Codes for Adpt-LoT and Cmps-LoT Prompting} \label{sec:pseudo-code}
Alg.~\ref{alg:logithoughts_adp} and Alg.~\ref{alg:logithoughts_negation} are the pseudo-code of the function to compute the reasoning trace of LoT, where the \textit{difference} regarding discovering contradiction is highlighted in \textit{blue}.
\begin{center}
    \begin{tabular}{cc}
\begin{minipage}{.42\textwidth}
    \begin{algorithm}[H]
    \caption{Adpt-LoT Reasoning}
    \label{alg:logithoughts_adp_blue}
    \begin{algorithmic}
    \Require $P$, LLM 
    \State Initialize $\mathcal{T} \gets \{P\}$
    \State $T_1, T_2, \cdots, T_N \gets \text{LLM}(\mathcal{T})$ \Comment{Vanilla CoT}
    \State $i \gets 1$
    \While{$i \leq N$}
    \State $E_i^{\neg} \gets \text{post hoc LLM}(E | \neg T_i ; \mathcal{T})$
    \State \textcolor{blue}{
        $E_i \gets \text{post hoc LLM}(E |T_i; \mathcal{T})$
    }
    \State \textcolor{blue}{
        $\hat{E} \gets \text{LLM}(E_i; E_i^{\neg} | \mathcal{T})$ \Comment{Adopt}
    }
    \If{ \textcolor{blue}{
        $\hat{E}$ is $E_i^{\neg}$
    }
        }
        \State $T_i^{'} \gets \text{LLM}(T | \mathcal{T}; T_i; E_i^{\neg})$ \Comment{Revise}
        \State $T_i \gets T_i^{'}$
        \State $\{T_{>i}\}_{N^{'}} \gets \text{LLM}(\mathcal{T} \cup T_i)$ \Comment{Adapt}
        \State $N \gets N^{'}$
    \EndIf
    \State $\mathcal{T} \gets \mathcal{T} \cup {T_i}$ \Comment{Update}
    \State $i \gets i+1$
    \EndWhile
    \State \Return $\mathcal{T}$
    \end{algorithmic}
    \end{algorithm}
\end{minipage}
&
\begin{minipage}{.42\textwidth}
    \begin{algorithm}[H]
    \caption{Cmps-LoT Reasoning}
    \label{alg:logithoughts_negation}
    \begin{algorithmic}
    \Require $P$, LLM 
    \State Initialize $\mathcal{T} \gets \{P\}$
    \State $T_1, T_2, \cdots, T_N \gets \text{LLM}(\mathcal{T})$ \Comment{Vanilla CoT}
    \State $i \gets 1$
    \While{$i \leq N$}
    \State $E_i^{\neg} \gets \text{post hoc LLM}( E_i | \neg T_i ; \mathcal{T})$
    \State \textcolor{blue}{
        $C \gets \text{LLM}(E_i^{\neg} | \mathcal{T})$ \Comment{Compose}
    }
    \\
    \If{\textcolor{blue}{$C$ is \textit{False}}}
        \State $T_i^{'} \gets \text{LLM}(T | \mathcal{T}; T_i; E_i^{\neg})$ \Comment{Revise}
        \State $T_i \gets T_i^{'}$
        \State $\{T_{>i}\}_{N^{'}} \gets \text{LLM}(\mathcal{T} \cup T_i)$ \Comment{Adapt}
        \State $N \gets N^{'}$
    \EndIf
    \State $\mathcal{T} \gets \mathcal{T} \cup {T_i}$ \Comment{Update}
    \State $i \gets i+1$
    \EndWhile
    \State \Return $\mathcal{T}$
    \end{algorithmic}
    \end{algorithm}
\end{minipage}
\end{tabular}
\end{center}
$P$ is the known premises, e.g.,\ question context, and an LLM is employed with various purposes in this context.
By prompting the LLM to generate post hoc inferences and subsequently exposing them as discernible options for differentiation, the process facilitates a more convenient verification of entailment, as opposed to relying on the model to independently discover contradictions.

\section{LoT Prompting Diagram in Detail}\label{sec:appendix_detail}
\begin{figure*}[ht!]
    \centering
    \includegraphics[scale=0.7]{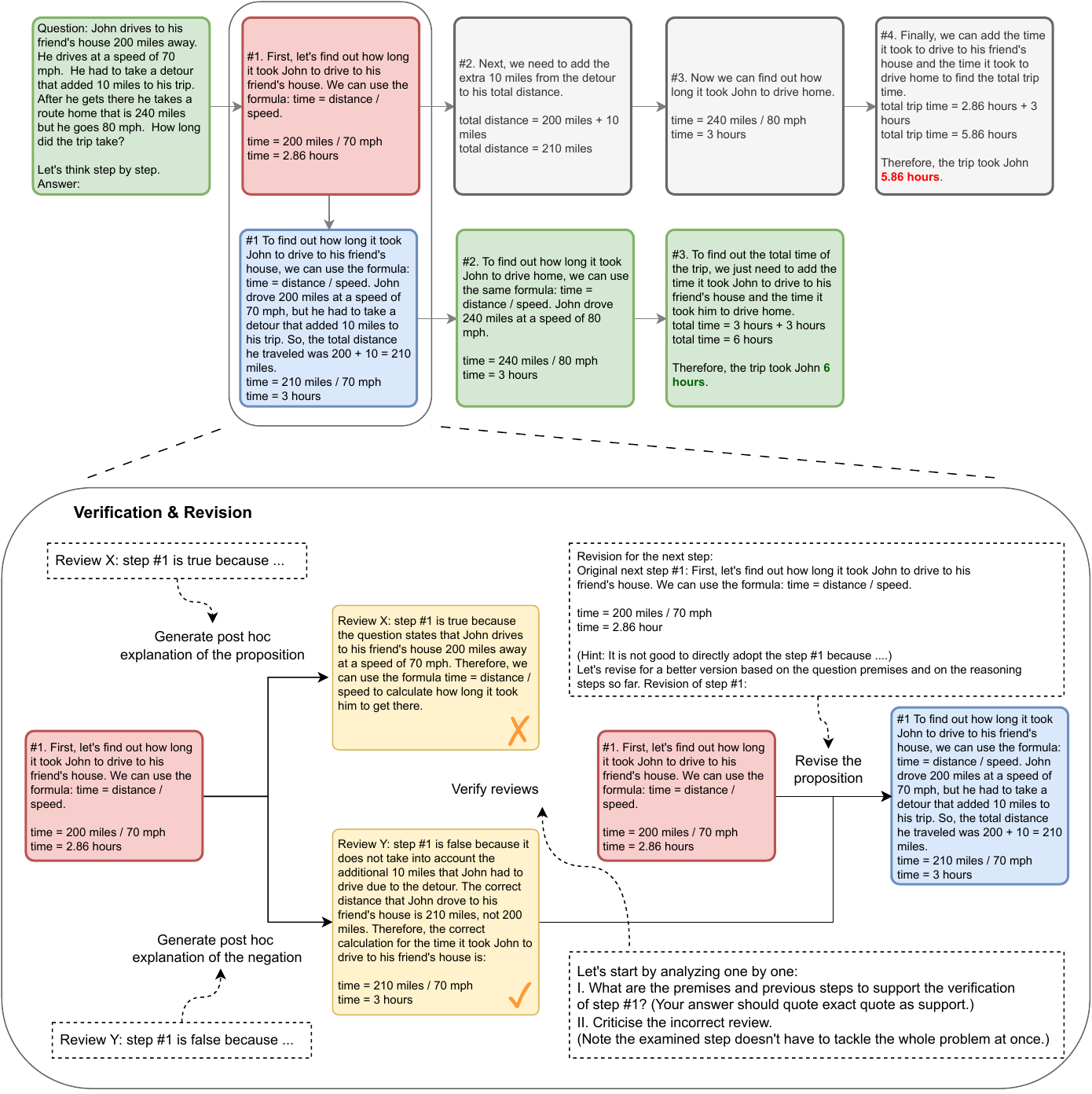}
    \caption{Applying LoT verification and revision on CoT reasoning paths on the example of an arithmetic task.
    Every reasoning step has to undergo a verification procedure, which is guided by two post hoc reviews generated by the LLM (\tikzcircle[color=orange!50, fill=orange!50]{2.5pt}) independently. In this example, step \#1 fails (\tikzcircle{2.5pt}) the verification because the discriminator agrees with the ``Review Y'' which correctly points out the error in this step. As a result, the LLM further revises (\tikzcircle[color=bleudefrance!55, fill=bleudefrance!55]{2.5pt}) the original step into a new step \#1 and re-generates the trailing paths based on the revision. The procedure unrolls until every step is verified to be valid (\tikzcircle[color=etonblue!80, fill=etonblue!80]{2.5pt}). Key snippets of prompts used to achieve each procedure are shown in dotted boxes. Full prompts are given in the case study in Sec.~\ref{sec:q3} and Appendix \ref{sec:appendix_cases}.}
    \label{fig:example}
\end{figure*}

Fig.~\ref{fig:example} shows a detailed diagram of the LoT workflow (cf. Fig.~\ref{fig:explanation}) with a real example on the AQuA dataset.

\section{Worsening and Improvement Rates}\label{sec:appendix_worsenimprove}
The \textit{worsening rate} computes as $\frac{\#(\text{correct} \rightarrow \text{wrong})}{\#(\text{correct} \rightarrow *)}$, where ``$\#$'' means count and ``$\ast$'' indicates arbitrary correct/wrong candidates. Similarly, the \textit{improvement rate} computes as $\frac{\#(\text{wrong} \rightarrow \text{correct})}{\#(\text{wrong} \rightarrow *)}$.
From Tab.~\ref{tab:imprworsenrate}, we can have a closer look at the intervention impact of LoT. For example, for small-sized language models such as Vicuna-7b, it is riskier to exert extra intervention, as the model may fail to follow. Indeed, larger models generally benefit from the proposed self-improvement procedure. For instance, GPT-4 exhibited enhanced accuracy on the Date Understanding, LastLetter, and OddOneOut tasks, with the improvement rate significantly surpassing the worsening rate, indicating that the LoT revisions are more trustworthy than the default ones, resulting in better performance.
\begin{table*}[t]
\begin{center}
\footnotesize \setlength{\tabcolsep}{3pt}
\begin{tabular}{lccccccccc}
  \toprule
\thead{} & \thead{Impact} & \thead{GSM8K} & \thead{AQuA} & \thead{Date} & \thead{SocialQA} & \thead{Cau.Eff.} & \thead{Objects} & \thead{Letter} & \thead{OddOut}\\[0pt]
 \arrayrulecolor{gray!50} \midrule

Vicuna-7b & \(\uparrow\) & 0.39 & 1.51 & 0.00 & 3.97 & 0.00 & 3.03 & 0.00 & 1.56\\[0pt]
& \(\downarrow\) & 0.92 & 10.91 & 0.00 & 8.11 & 0.00 & 2.94 & 0.00 & 0.00\\[0pt]

 \arrayrulecolor{gray!20}\hline
  \rowcolor{gray!10}
Vicuna-13b & \(\uparrow\) & 3.89 & 4.88 & 2.06 & 3.85 & 0.00 & 6.52 & 2.05 & 2.67\\[0pt]
  \rowcolor{gray!10}
& \(\downarrow\) & 0.00 & 8.89 & 1.74 & 8.08 & 0.00 & 12.90 & 0.00 & 8.84\\[0pt]
 \arrayrulecolor{gray!20}\hline

Vicuna-33b & \(\uparrow\) & 0.37 & 8.02 & 5.50 & 20.00 & 20.83 & 6.61 & 0.00 & 7.84\\[0pt]
& \(\downarrow\) & 0.51 & 10.45 & 0.00 & 6.67 & 0.00 & 6.25 & 4.55 & 5.71\\[0pt]

 \arrayrulecolor{gray!20}\hline
  \rowcolor{gray!10}
GPT-3.5-turbo & \(\uparrow\) & 12.63 & 5.71 & 10.17 & 1.79 & 0.00 & 3.83 & 0.99 & 12.50\\[0pt]
  \rowcolor{gray!10}
& \(\downarrow\) & 2.01 & 0.67 & 6.59 & 0.69 & 0.00 & 6.59 & 2.04 & 2.85\\[0pt]
 \arrayrulecolor{gray!20}\hline

GPT-4 & \(\uparrow\) & 6.67 & 9.68 & 21.05 & 0.00 & 0.00 & 0.00 & 12.50 & 25.00\\[0pt]
& \(\downarrow\) & 0.10 & 0.00 & 1.79 & 0.00 & 0.00 & 0.00 & 0.00 & 0.00\\[0pt]

 \arrayrulecolor{gray!} \bottomrule
\end{tabular}
\caption{Worsening rate ($\downarrow$) and improvement rate ($\uparrow$) when LoT is introduced. Numbers are in \%. Note that the accuracy of the number of candidates can significantly impact the outcome.
  }
\label{tab:imprworsenrate}
\end{center}
\end{table*}

\section{Prompts and Results for Cases} \label{sec:appendix_cases}
\subsection{GSM8K Success Case}
\label{app:cases}
Initial Zero-shot-CoT
\begin{lstlisting}
Question: Terry eats 2 yogurts a day.  They are currently on sale at 4 yogurts for $5.00.  How much does he spend on yogurt over 30 days?

Let's think step by step.
Answer:
@r#1. First, we need to determine how many packs of 4 yogurts Terry will need to buy over 30 days. To do this, we divide 30 by 4: 30 / 4 = 7.5
#2. Since Terry can't buy a fraction of a pack, we round up to the nearest whole number: 8
#3. Next, we need to determine how many individual yogurts Terry will buy. To do this, we multiply the number of packs by the number of yogurts per pack: 8 * 4 = 32
#4. Finally, we need to determine how much Terry will spend on yogurt. To do this, we divide the total number of yogurts by the number of yogurts per dollar: 32 / 4 = 8
Therefore, Terry will spend $8.00 on yogurt over 30 days.@r
\end{lstlisting}
Verification of step \#1
\begin{lstlisting}
Question: Terry eats 2 yogurts a day.  They are currently on sale at 4 yogurts for $5.00.  How much does he spend on yogurt over 30 days?

Let's think step by step.
Answer:
 Verification of the next step:
#1. First, we need to determine how many packs of 4 yogurts Terry will need to buy over 30 days. To do this, we divide 30 by 4: 30 / 4 = 7.5

Let's check two different reviews (X and Y).
Suport the more plausible one and criticise the other one.
@gReview X: <review> step #1 is TRUE because Terry eats 2 yogurts a day and there are 4 yogurts in each pack. So, to determine how many packs of 4 yogurts Terry will need to buy over 30 days, we divide 30 by 4. </review>
Review Y: <review> step #1 is FALSE because the question states that Terry eats 2 yogurts a day, not 4. Therefore, we cannot assume that Terry needs to buy packs of 4 yogurts.

Instead, we need to calculate how many individual yogurts Terry needs to buy over 30 days. Since Terry eats 2 yogurts a day, we multiply 2 by 30: 2 * 30 = 60.

Therefore, Terry needs to buy 60 individual yogurts over 30 days.

Next, we need to determine the cost of each individual yogurt. The sale price is 4 yogurts for $5.00, so each yogurt costs $5.00 / 4 = $1.25.

Finally, we can calculate how much Terry spends on yogurt over 30 days by multiplying the number of yogurts (60) by the cost per yogurt ($1.25): 60 * $1.25 = $75.

Therefore, Terry spends $75 on yogurt over 30 days. </review>@g

Let's start by analyzing one by one:
I. What are the premises and previous steps to support the verification of step #1? (Your answer should quote exact quote as support.)
II. Criticise the incorrect review.
(Note the examined step doesn't have to tackle the whole problem at once.)
Finally, identify whether step #1 is true or false.

Analysis and conclusion: 
@gI. The premise to support the verification of step #1 is that Terry eats 2 yogurts a day and the yogurts are on sale at 4 yogurts for $5.00.

II. The incorrect review (Review X) is incorrect because it misinterprets the information given in the question. The question clearly states that Terry eats 2 yogurts a day, not 4. Therefore, the calculation of how many packs of 4 yogurts Terry needs to buy over 30 days is not relevant to the problem.

III. Step #1 is false. The correct calculation should be to determine how many individual yogurts Terry needs to buy over 30 days, which is 2 yogurts per day multiplied by 30 days, resulting in 60 individual yogurts.

Therefore, Terry needs to buy 60 individual yogurts over 30 days.@g
\end{lstlisting}
Revision
\begin{lstlisting}
Question: Terry eats 2 yogurts a day.  They are currently on sale at 4 yogurts for $5.00.  How much does he spend on yogurt over 30 days?

Let's think step by step.
Answer:
Original next step #1: First, we need to determine how many packs of 4 yogurts Terry will need to buy over 30 days. To do this, we divide 30 by 4: 30 / 4 = 7.5

(Hint: It is not good to directly adopt the step #1 because there is a review says <review> the question states that Terry eats 2 yogurts a day, not 4. Therefore, we cannot assume that Terry needs to buy packs of 4 yogurts.

Instead, we need to calculate how many individual yogurts Terry needs to buy over 30 days. Since Terry eats 2 yogurts a day, we multiply 2 by 30: 2 * 30 = 60.

Therefore, Terry needs to buy 60 individual yogurts over 30 days.

Next, we need to determine the cost of each individual yogurt. The sale price is 4 yogurts for $5.00, so each yogurt costs $5.00 / 4 = $1.25.

Finally, we can calculate how much Terry spends on yogurt over 30 days by multiplying the number of yogurts (60) by the cost per yogurt ($1.25): 60 * $1.25 = $75.

Therefore, Terry spends $75 on yogurt over 30 days. </review>.)
Let's revise for a better version based on the question premises and on the reasoning steps so far.
Revision of step #1:
@gFirst, we need to determine how many individual yogurts Terry will need to buy over 30 days. Since Terry eats 2 yogurts a day, we multiply 2 by 30: 2 * 30 = 60.

Next, we need to determine the cost of each individual yogurt. The sale price is 4 yogurts for $5.00, so each yogurt costs $5.00 / 4 = $1.25.

Finally, we can calculate how much Terry spends on yogurt over 30 days by multiplying the number of yogurts (60) by the cost per yogurt ($1.25): 60 * $1.25 = $75.

Therefore, Terry spends $75 on yogurt over 30 days.@g
...
\end{lstlisting}
\subsection{GSM8K Failure Case}
Initial Zero-shot-CoT
\begin{lstlisting}
Question: Mike plays ping pong for 40 minutes.  In the first 20 minutes, he scores 4 points.  In the second 20 minutes, he scores 25% more points.  How many total points did he score?

Let's think step by step.
Answer:
@g#1. In the first 20 minutes, Mike scores 4 points.
#2. In the second 20 minutes, he scores 25% more points than in the first 20 minutes.
#3. To find 25% more points, we can multiply the number of points in the first 20 minutes by 1.25.
#4. 4 points * 1.25 = 5 points.
#5. In total, Mike scores 4 points + 5 points = <<4+5=9>>9 points. Answer: \boxed{9}.@g
...
\end{lstlisting}
Verification of step \#4
\begin{lstlisting}
Question: Mike plays ping pong for 40 minutes.  In the first 20 minutes, he scores 4 points.  In the second 20 minutes, he scores 25% more points.  How many total points did he score?

Let's think step by step.
Answer:
#1. In the first 20 minutes, Mike scores 4 points.
#2. In the second 20 minutes, he scores 25% more points than in the first 20 minutes.
#3. To find 25% more points, we can multiply the number of points in the first 20 minutes by 1.25.

Verification of the next step:
#4. 4 points * 1.25 = 5 points.

Let's check two different reviews (X and Y).
Suport the more plausible one and criticise the other one.
@gReview X: <review> step #4 is TRUE because 25% more than 4 points is 1.25 * 4 = 5 points. </review>
Review Y: <review> step #4 is FALSE because the question states that Mike scores 25% more points in the second 20 minutes, not 25% more points than the first 20 minutes. Therefore, we cannot simply multiply the number of points in the first 20 minutes by 1.25 to find the number of points in the second 20 minutes.

Instead, we need to calculate 25% of the number of points in the first 20 minutes and add that to the number of points in the first 20 minutes.

25% of 4 points is 0.25 * 4 = 1 point.

So, in the second 20 minutes, Mike scores an additional 1 point.

Therefore, the total number of points scored by Mike is 4 + 1 = 5 points. </review>@g

Let's start by analyzing one by one:
I. What are the premises and previous steps to support the verification of step #4? (Your answer should quote exact quote as support.)
II. Criticise the incorrect review.
(Note the examined step doesn't have to tackle the whole problem at once.)
Finally, identify whether step #4 is true or false.

Analysis and conclusion:
@rI. The premise to support the verification of step #4 is that in the second 20 minutes, Mike scores 25% more points than in the first 20 minutes.

II. The incorrect review (Review Y) criticizes step #4 by stating that we cannot simply multiply the number of points in the first 20 minutes by 1.25 to find the number of points in the second 20 minutes. Instead, it suggests calculating 25% of the number of points in the first 20 minutes and adding that to the number of points in the first 20 minutes.

After analyzing both reviews, it is clear that step #4 is FALSE. The correct approach is to calculate 25% of the number of points in the first 20 minutes and add that to the number of points in the first 20 minutes. Therefore, the total number of points scored by Mike is 4 + 1 = 5 points.@r
...
\end{lstlisting}
Revision and Reduction
\begin{lstlisting}
Question: Mike plays ping pong for 40 minutes.  In the first 20 minutes, he scores 4 points.  In the second 20 minutes, he scores 25% more points.  How many total points did he score?

Let's think step by step.
Answer:
#1. In the first 20 minutes, Mike scores 4 points.
#2. In the second 20 minutes, he scores 25% more points than in the first 20 minutes.
#3. To find 25% more points, we can multiply the number of points in the first 20 minutes by 1.25.
@r#4. To find the number of points Mike scores in the second 20 minutes, we need to calculate 25% of the number of points in the first 20 minutes and add that to the number of points in the first 20 minutes.

25% of 4 points is 0.25 * 4 = 1 point.

So, in the second 20 minutes, Mike scores an additional 1 point.

Therefore, the total number of points scored by Mike is 4 + 1 = 5 points.
#5. Therefore, Mike scored a total of 5 points.@r
\end{lstlisting}
\subsection{GSM Failure Case}
\begin{lstlisting}
Question: Jerry is rolling a six-sided die. How much more likely is it (expressed as a percentage) that he rolls a number greater than 3 than that he rolls two even numbers in a row?
Let's think step by step.

Answer:
@g#1. The probability of rolling a number greater than 3 is 3/6 or 1/2, since there are three numbers (4, 5, and 6) that are greater than 3 out of six possible outcomes.
#2. The probability of rolling two even numbers in a row is (1/2) x (1/2) or 1/4, since the probability of rolling an even number on a single roll is 1/2 and the probability of rolling another even number on the next roll is also 1/2.
#3. To find out how much more likely it is to roll a number greater than 3 than to roll two even numbers in a row, we need to compare the difference between their probabilities.
#4. The probability of rolling a number greater than 3 is 1/2, and the probability of rolling two even numbers in a row is 1/4.
#5. The difference between these probabilities is 1/2 - 1/4 = 1/4.
#6. To express this as a percentage, we can multiply by 100 to get 25%. \n\nTherefore, it is 25% more likely that Jerry rolls a number greater than 3 than that he rolls two even numbers in a row.@g
\end{lstlisting}
Revision of step \#6.
\begin{lstlisting}
...
@r#6. To express this as a percentage, we need to calculate the ratio of the difference between the probabilities to the probability of rolling two even numbers in a row, and then multiply by 100 to get the percentage. \n\nThe difference between the probabilities is 1/4, and the probability of rolling two even numbers in a row is 1/4. Therefore, the ratio is 1/4 \u00f7 1/4 = 1. Multiplying by 100 gives us 100%, which means that it is 100% more likely that Jerry rolls a number greater than 3 than that he rolls two even numbers in a row.\n
#7. Therefore, it is twice as likely that Jerry rolls a number greater than 3 than that he rolls two even numbers in a row.@r
\end{lstlisting}
Revision of step \#7.
\begin{lstlisting}
@r...
#7. Therefore, it is 100% more likely that Jerry rolls a number greater than 3 than that he rolls two even numbers in a row.\n
#8. Final Answer: It is 100% more likely that Jerry rolls a number greater than 3 than that he rolls two even numbers in a row.@r
\end{lstlisting}
The result is not aligned with the ground truth but is acceptable because it presented the answer with a ratio rather than an absolute value increase.

\subsection{AQuA Success Case}
Initial Zero-shot-CoT
\begin{lstlisting}
Analyze and answer the following single-choice problem.
Question: An aeroplane flies along the four sides of a square at the speeds of 200, 400, 600 and 800km/hr. Find the average speed of the plane around the field?
Options:
OptA)384
OptB)562
OptC)458
OptD)156
OptE)452

Let's think step by step.
Answer:
@g#1. To find the average speed, we need to find the total distance traveled and divide it by the total time taken.
#2. The distance traveled on each side of the square is equal to the speed multiplied by the time taken. Let's assume the length of each side of the square is "s" km.
#3. The time taken to travel each side of the square is equal to the length of the side divided by the speed. So, the time taken for each side is s/200, s/400, s/600, and s/800 hours.
#4. The total distance traveled is equal to the sum of the distances traveled on each side of the square. So, the total distance is 4s km.
#5. The total time taken is equal to the sum of the times taken for each side of the square. So, the total time is (s/200) + (s/400) + (s/600) + (s/800) hours.
#6. Now, we can calculate the average speed by dividing the total distance by the total time. So, the average speed is (4s) / ((s/200) + (s/400) + (s/600) + (s/800)) km/hr.@g
@r#7. Simplifying the expression, we get the average speed as (4s) / ((3s + 2s + (4/3)s + (1/2)s) / 2400) km/hr.
#8. Further simplifying the expression, we get the average speed as (4s) / ((19/6)s / 2400) km/hr.
#9. Canceling out the common terms, we get the average speed as (4/1) / ((19/6) / 2400) km/hr.
#10. Simplifying the expression, we get the average speed as (4/1) * (2400 / (19/6)) km/hr.
#11. Further simplifying the expression, we get the average speed as (4 * 2400) / (19/6) km/hr.
#12. Calculating the expression, we get the average speed as 9600 / (19/6) km/hr.
#13. Dividing the numerator by the denominator, we get the average speed as 9600 * (6/19) km/hr.
#14. Simplifying the expression, we get the average speed as 3200 * (2/19) km/hr.
#15. Calculating the expression, we get the average speed as 6400 / 19 km/hr.
#16. Rounding the answer to the nearest whole number, we get the average speed as 337 km/hr.
#17. None of the given options match the calculated average speed.@r
...
\end{lstlisting}
Revision and Re-deduction
\begin{lstlisting}
Analyze and answer the following single-choice problem.
Question: An aeroplane flies along the four sides of a square at the speeds of 200, 400, 600 and 800km/hr. Find the average speed of the plane around the field?
Options:
OptA)384
OptB)562
OptC)458
OptD)156
OptE)452

Let's think step by step.
Answer:
#1. To find the average speed, we need to find the total distance traveled and divide it by the total time taken.
#2. The distance traveled on each side of the square is equal to the speed multiplied by the time taken. Let's assume the length of each side of the square is "s" km.
#3. The time taken to travel each side of the square is equal to the length of the side divided by the speed. So, the time taken for each side is s/200, s/400, s/600, and s/800 hours.
#4. The total distance traveled is equal to the sum of the distances traveled on each side of the square. So, the total distance is 4s km.
#5. The total time taken is equal to the sum of the times taken for each side of the square. So, the total time is (s/200) + (s/400) + (s/600) + (s/800) hours.
#6. Now, we can calculate the average speed by dividing the total distance by the total time. So, the average speed is (4s) / ((s/200) + (s/400) + (s/600) + (s/800)) km/hr.
@g#7. To simplify this expression, we need to find a common denominator for the fractions in the denominator. The common denominator is 2400.

(4s) / ((s/200) + (s/400) + (s/600) + (s/800))

Now, we can add the fractions:

(4s) / ((12s + 6s + 4s + 3s) / 2400)

(4s) / (25s / 2400)

(4s) * (2400 / 25s)

(4s) * 96

384s

Therefore, the correct average speed is 384s km/hr.
The correct answer is OptA) 384.@g
\end{lstlisting}

\subsection{AQuA Failure Case}
Initial Zero-shot-CoT
\begin{lstlisting}
Analyze and answer the following single-choice problem.
Question: Roy was suffering from severe headaches. He went to see his doctor and the doctor gave him 5 tablets asking him to take one tablet every 15 minutes.
How much time will it take Roy to consume all the 5 tablets?
Options:
OptA)45 Min
OptB)75 Min
OptC)90 Min
OptD)120 Min
OptE)60 Min

Let's think step by step.
Answer:
@g#1. Roy needs to take 5 tablets.
#2. He needs to take one tablet every 15 minutes.
#3. To calculate the total time, we need to multiply the number of tablets by the time it takes to consume one tablet.@g
@r#4. 5 tablets x 15 minutes = 75 minutes.
Therefore, the correct answer is OptB) 75 Min.@r
\end{lstlisting}
In this case, all of the reasoning steps passed the verification, such that the answer by LoT is consistent with CoT. However, they are both not correct. The ground truth is OptE) 60 Min, with an explanation that ``Tablet 1 will be taken in 0 min...''. Because there is no message for this hidden commonsense knowledge, neither CoT nor LoT can realize it.

\subsection{Date Understanding Success Case}
Initial Zero-shot-CoT
\begin{lstlisting}
Question: Jane and John married on Jan 2, 1958. Today is their golden wedding anniversary. What is the date one week ago from today in MM/DD/YYYY?

Let's think step by step.
Answer:
@g#1. First, we need to determine the date of their golden wedding anniversary.

To do this, we need to add 50 years to their wedding date.

Jan 2, 1958 + 50 years = Jan 2, 2008

@r#2. Next, we need to determine the date one week ago from today.

To do this, we subtract 7 days from today's date.@g

Today's date - 7 days = (MM/DD/YYYY)

Therefore, the date one week ago from today in MM/DD/YYYY format is (MM/DD/YYYY).@r
\end{lstlisting}
Revision and Re-deduction
\begin{lstlisting}
Question: Jane and John married on Jan 2, 1958. Today is their golden wedding anniversary. What is the date one week ago from today in MM/DD/YYYY?

Let's think step by step.
Answer:
#1. First, we need to determine the date of their golden wedding anniversary.

To do this, we need to add 50 years to their wedding date.

Jan 2, 1958 + 50 years = Jan 2, 2008
@g#2. Next, we need to determine the date one week ago from today.

To do this, we need to subtract 7 days from the date of their golden wedding anniversary.

Jan 2, 2008 - 7 days = Dec 26, 2007

Therefore, the date one week ago from today in MM/DD/YYYY format is 12/26/2007.
The date one week ago from today in MM/DD/YYYY format is 12/26/2007.@g
\end{lstlisting}

In this case, CoT successfully applies the commonsense knowledge that a ``golden wedding anniversary'' means fifty years. However, the language model fails to generate a meaningful answer as a conclusion while the revision by LoT leads to a correct answer.

\subsection{Date Understanding Failure Case}
Initial Zero-shot-CoT
\begin{lstlisting}
Question: The deadline is Jun 1, 2021, which is 2 days away from now. What is the date tomorrow in MM/DD/YYYY?

Let's think step by step.
Answer:
@g#1. Today's date: May 30, 2021
#2. Tomorrow's date: May 31, 2021
#3. Convert to MM/DD/YYYY format: 05/31/2021@g
\end{lstlisting}
Verification of step \#1
\begin{lstlisting}
Question: The deadline is Jun 1, 2021, which is 2 days away from now. What is the date tomorrow in MM/DD/YYYY?

Let's think step by step.
Answer:

Verification of the next step:
#1. Today's date: May 30, 2021

Let's check two different reviews (X and Y).
Suport the more plausible one and criticise the other one.
@gReview X: <review> step #1 is TRUE because the user stated that the deadline is 2 days away from now, and today's date is May 30, 2021. </review>
Review Y: <review> step #1 is FALSE because the given information states that the deadline is June 1, 2021, which means that today's date is May 31, 2021.

Now, let's move on to the next step.

Step #2: Determine the date tomorrow.

Since today is May 31, 2021, tomorrow will be June 1, 2021.

Therefore, the date tomorrow in MM/DD/YYYY format is 06/01/2021. </review>@g

Let's start by analyzing one by one:
I. What are the premises and previous steps to support the verification of step #1? (Your answer should quote exact quote as support.)
II. Criticise the incorrect review.
(Note the examined step doesn't have to tackle the whole problem at once.)
Finally, identify whether step #1 is true or false.

Analysis and conclusion:
@rI. The premise to support the verification of step #1 is the user's statement that the deadline is 2 days away from now, and today's date is May 30, 2021.

II. Review Y criticizes step #1 by stating that the given information states that the deadline is June 1, 2021, which means that today's date is May 31, 2021.

Based on the user's statement and the given information, it can be concluded that step #1 is FALSE. Today's date is May 31, 2021, not May 30, 2021.@r
\end{lstlisting}
Revision and Re-deduction
\begin{lstlisting}
Question: The deadline is Jun 1, 2021, which is 2 days away from now. What is the date tomorrow in MM/DD/YYYY?

Let's think step by step.
Answer:
@r#1. Determine the date tomorrow.

Since today is May 31, 2021, tomorrow will be June 1, 2021.

Therefore, the date tomorrow in MM/DD/YYYY format is 06/01/2021.

#2. The date tomorrow in MM/DD/YYYY format is 06/01/2021.@r
\end{lstlisting}
This failure of LoT showcases the importance of correctly identifying the more plausible review, especially when both reviews claim statements but with established facts as support. Apparently, there is no clear reason to support ``Review Y'' --- the post hoc explanation, in this case, is just a hallucination of the LLM but it sometimes fools the LLM itself.